\documentclass{article}

\PassOptionsToPackage{numbers}{natbib}
\usepackage[sglblindworkshop,final]{neurips_2025}

\workshoptitle{NeurIPS 2025 AI for Science workshop}

\makeatletter
\renewcommand{\@noticestring}{%
  NeurIPS 2025 AI for Science workshop.%
}
\makeatother

\usepackage[utf8]{inputenc} % allow utf-8 input
\usepackage[T1]{fontenc}    % use 8-bit T1 fonts
\usepackage{hyperref}       % hyperlinks
\usepackage{url}            % simple URL typesetting
\usepackage{booktabs}       % professional-quality tables
\usepackage{amsfonts}       % blackboard math symbols
\usepackage{amsmath}
\usepackage{nicefrac}       % compact symbols for 1/2, etc.

\usepackage{tikz}
\usetikzlibrary{positioning} % needed for 'of' syntax
\usetikzlibrary{shapes.geometric}
\usetikzlibrary{calc}
\usetikzlibrary{arrows.meta}
\usepackage{microtype}      % microtypography
\usepackage{xcolor}         % colors
\usepackage{multirow}

\title{Geometry Aware Inference of Steady State PDEs Using Equivariant Neural Field Representations.}

\author{
  Giovanni Catalani \\
  Airbus, ISAE-Supaero \\
  Toulouse, France \\
  \texttt{giovanni.catalani@airbus.com}
  \And
  Michael Bauerheim \\
  ISAE-Supaero \\
  Toulouse, France \\
  \texttt{michael.bauerheim@isae-supaero.fr}
  \And
  Frédéric Tost \\
  Airbus \\
  Toulouse, France \\
  \texttt{frederic.tost@airbus.com}
  \And
  Xavier Bertrand \\
  Airbus \\
  Toulouse, France \\
  \texttt{xavier.bertrand@airbus.com}
  \And
  Joseph Morlier \\
  ISAE-Supaero, ICA, Université de Toulouse \\
  Toulouse, France \\
  \texttt{joseph.morlier@isae-supaero.fr}
}

\begin{document}

\maketitle

\begin{abstract}
Advances in neural operators have introduced discretization invariant surrogate models for PDEs on general geometries, yet many approaches struggle to encode local geometric structure and variable domains efficiently. We introduce enf2enf, a neural field approach for predicting steady-state PDEs with geometric variability. 
Our method encodes geometries into latent features anchored at specific spatial locations, preserving locality throughout the network. These local representations are combined with global parameters and decoded to continuous physical fields, enabling effective modeling of complex shape variations.
Experiments on aerodynamic and structural benchmarks demonstrate competitive or superior performance compared to graph-based, neural operator, and recent neural field methods, with real-time inference and efficient scaling to high-resolution meshes.
\end{abstract}

\section{Introduction}
Partial Differential Equations (PDEs) are fundamental to modeling physical phenomena across numerous scientific and engineering domains \cite{debnath2005nonlinear,temam2024navier}. Computational Fluid Dynamics (CFD), for instance,  has become an indispensable tool for aircraft and vehicle design, enabling high-fidelity simulations of complex aerodynamic phenomena. However, the computational cost of traditional numerical methods remains a significant bottleneck \cite{ferziger2019computational}. The complexity arises from the large number of degrees of freedom required to capture multiscale physics, mesh refinement requirements for accurate boundary layer resolution, and iterative solution procedures. The need for computationally efficient simulations and the abundance of available data motivate the development of data-driven PDE surrogate models that can offer remarkable speedups compared to traditional numerical methods, while reducing the level of fidelity \cite{brunton2020machine}.\\
In this context, deep neural networks have complemented existing reduced order models based on linear dimensionality reduction and interpolation \cite{berkooz1993proper}, which are typically limited to handle fixed geometries, providing significant advantages for physical modeling tasks with geometric variability.
Early neural network architectures for PDE modeling exploited spatial inductive biases through Convolutional Neural Networks (CNNs) \cite{ajuria2020towards} on regular grids or Graph Neural Networks (GNNs) \cite{pfaff2020learning} on irregular meshes. However, CNNs necessitate costly interpolation procedures when handling unstructured computational grids \cite{catalani2023comparative}, while many GNN-based approaches encounter significant computational bottlenecks on large-scale industrial meshes \cite{hines2023graph} and demonstrate poor generalization across different mesh resolutions \cite{fortunato2022multiscale}. To overcome these discretization-dependent limitations, Neural operators have emerged as a promising alternative by framing the problem as learning mappings between infinite-dimensional function spaces rather than learning input-output relationships defined on fixed resolutions or specific discretizations. This paradigm enables discretization-invariant learning across different mesh resolutions. Notable examples include DeepONet \cite{lu2019deeponet} and Fourier Neural Operators (FNO) \cite{li2020fourier}, which have shown impressive performance across various PDE benchmarks. Nevertheless, these methods suffer from architectural constraints: DeepONet requires manual design of input function encodings, while FNO is fundamentally limited to structured grids \cite{li2023fourier} and periodic boundary conditions, hindering their deployment in complex geometric scenarios. More recent approaches based on Neural Fields, such as CORAL \cite{serrano2024operator} address these limitations by encoding input functions into compact global latent representations using implicit neural networks. However, these approaches rely on global latent representations that struggle to preserve local geometric structure. This limitation is particularly problematic for PDEs on variable geometries, where spatially localized physical phenomena are crucial for accurate modeling. The idea of using spatially grounded latent representation has been recently introduced for dynamics modeling on fixed domains \cite{knigge2024space} using Equivariant Neural Field. However, the challenge of geometry-aware inference for steady-state PDEs across changing geometries has not been tackled, despite its paramount importance in practical engineering scenarios.
\paragraph{Contribution.}To address these limitations, we propose \texttt{enf2enf}, a novel geometry-aware operator learning framework specifically designed for steady-state PDEs on variable geometries. Our approach employs a geometry encoder that extracts spatially-grounded point cloud embeddings from input geometric fields (such as Signed Distance Functions), such that each latent point is anchored to a specific spatial location and encodes local geometric features within its neighborhood.
The decoder then processes these local geometric encodings by combining them with global operating conditions (such as inflow parameters). Self-attention mechanisms facilitate message passing between latent points, enabling spatial communication.
This encoder-decoder design bypasses the encode-process-decode pipeline employed in previous approaches \cite{serrano2024operator}, eliminating the intermediate latent space processor that introduces auxiliary objectives decoupled from the final PDE reconstruction error. 
The experimental validation focuses on applied PDE benchmarks, including structural modeling and large-scale aerodynamic simulations.

\section{Related Work}
\paragraph{Mesh-based surrogate models.} Traditional surrogate modeling approaches for PDEs rely on modal decomposition techniques such as Proper Orthogonal Decomposition (POD) \cite{berkooz1993proper} combined with Gaussian Process Regression (GPR) for interpolation \cite{saves2024smt}. While effective on fixed geometries, extending these methods to handle geometric variability remains challenging. Mesh morphing variants (MMGP) \cite{casenave2024mmgp} attempt to address this through reference mesh deformation but are constrained to similar topologies. Graph Neural Networks (GNNs), particularly MeshGraphNets \cite{pfaff2020learning}, emerged as alternatives for irregular meshes but face computational bottlenecks on large-scale problems and exhibit poor generalization across different mesh resolutions \cite{fortunato2022multiscale}.

\paragraph{Neural operators.}Operator learning addresses these limitations by modeling mappings between infinite-dimensional function spaces rather than discrete mesh representations. DeepONet \cite{lu2019deeponet} enables querying at arbitrary coordinates but requires input functions to be observed on predefined grids, necessitating identical observation patterns across training and testing. Fourier Neural Operators (FNO) \cite{li2020fourier} leverage fast Fourier transforms for efficient spectral domain computations but are fundamentally restricted to structured grids. Extensions like Geo-FNO \cite{li2023fourier} and Geometry-Informed Neural Operators (GINO) attempt to handle irregular geometries but do not show remarkable gains in performance, while lacking explicit mechanisms to encode geometric variability. Recent transformer-based approaches like Transolver \cite{wu2024transolver} incorporate attention mechanisms for spatial reasoning, and are able to locally extract slices, thereby improving accuracy for the surrogate modeling task on variable geometries.

\paragraph{Neural fields or Implicit Neural Representations.} Implicit Neural Representations (INRs) model spatial data as continuous functions \cite{tancik2020fourier,sitzmann2020implicit}, enabling coordinate-based querying at arbitrary locations. CORAL \cite{serrano2024operator} employs meta-learning to encode input functions into global latent vectors, then processes these representations in latent space before decoding to output fields through an encode-process-decode pipeline. However, compressing entire geometric configurations into single global vectors discards spatial locality and geometric structure. Equivariant Neural Fields (ENFs) \cite{wessels2024grounding} address this limitation by introducing geometrically grounded latent representations, but existing applications focus exclusively on time-dependent PDE modeling with fixed geometries \cite{knigge2024space}, still employing an intermediate processor network and without incorporation of global operating conditions. 
In contrast, our approach employs ENFs solely as geometry encoders, eliminates the intermediate processor, and directly combines local geometric conditioning with global operating parameters within the decoder network.

\section{Methodology}

\begin{figure}[h]
  \centering
  \begin{tikzpicture}
      % Place the original PDF
      \node[anchor=south west,inner sep=0] (image) at (0,0) {
          \includegraphics[width=\textwidth]{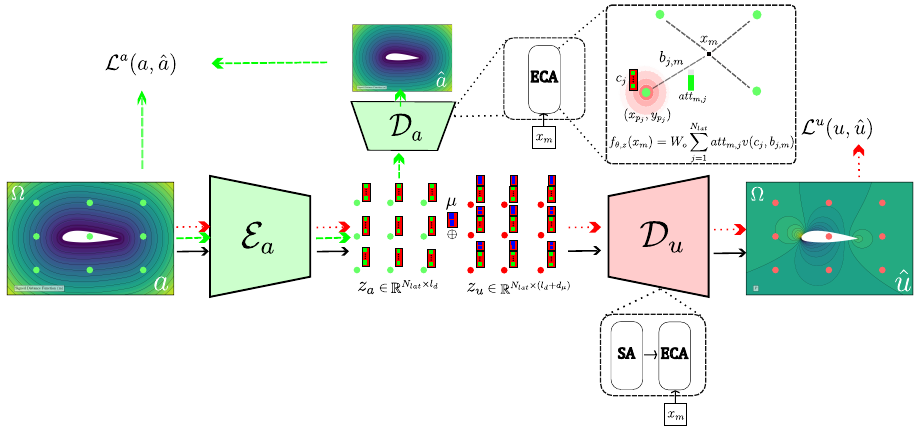}
      };
      
      % Get the size of the image to place legend appropriately
      \begin{scope}[x={(image.south east)},y={(image.north west)}]
          % Legend box
          \node[draw, fill=white, opacity=0.9, anchor=south west, font=\small] 
               at (0.01,0.1) {
              \begin{tabular}{l@{\hspace{0.05cm}}l}
                  \tikz\draw[dashed,green, line width=1pt] (0,0) -- (0.04,0); & \small{Input Enc. Training} \\
                  \tikz\draw[dotted,red,line width=1pt] (0,0) -- (0.04,0); & \small{Output Dec. Training} \\
                  \tikz\draw[line width=1pt] (0,0) -- (0.04,0); & \small{Inference}
              \end{tabular}
          };
      \end{scope}
  \end{tikzpicture}
  \caption{%
  \textbf{Overview of the \texttt{enf2enf} architecture.}
The input geometry field $a_i$ is encoded by $\mathcal{E}_{\mathrm{a}}$ into a \emph{spatially anchored} latent point cloud $z_a=\{(p_j,c_j^{\text{in}})\}_{j=1}^{N_{\text{lat}}}$ (green).
Each latent feature $c_j$ is node-wise concatenated with the global parameters $\mu$
to yield the conditioned set $z_u$ (red), which the decoder $\mathcal{D}_{\mathrm{u}}$ maps to the predicted physical field $\hat u_i$ via a composition of latent self attention (SA) and equivariant cross attention (ECA) decoding. The different training and inference paths are highlighted with the respective losses.%
  }
  \label{fig:improved_architecture}
\end{figure}

\subsection{Problem Formulation}

We consider the operator learning problem for steady state PDEs on variable geometries. Given input functions $a_i$ consisting of geometric descriptors (e.g., signed distance functions) and global parameters $\mu_i$ (e.g., inflow conditions, material properties), we seek to learn the mapping to output physical fields $u_i$ (e.g., pressure, stress) that satisfy the governing PDEs.

Formally, let $\Omega \subset \mathbb{R}^d$ be a bounded spatial domain where $d \in \{2,3\}$. We define the input space as $\mathcal{A} = \mathcal{A}_{\text{geom}}(\Omega) \times \mathcal{A}_{\mu}$, where $a_{\text{geom}}: \Omega \to \mathbb{R}^{n_a}$ encodes the geometry as functions over $\Omega$ (with $n_a$ denoting the input field dimension, e.g., $n_a=1$ for scalar SDF) and $\mu \in \mathcal{A}_{\mu} \subset \mathbb{R}^{l_\mu}$ represents global parameters. The governing PDEs define a nonlinear operator $\mathcal{G}^\dagger: \mathcal{A} \rightarrow \mathcal{U}(\Omega)$ mapping to the solution function space $\mathcal{U}(\Omega)$, where $u: \Omega \to \mathbb{R}^{n_u}$ (with $n_u$ denoting the output field dimension).

We approximate $\mathcal{G}^\dagger$ with a parameterized operator $\mathcal{G}_\theta$ learned from training pairs $(a_i, u_i)$ observed on unstructured meshes $\mathcal{X}_i \subset \Omega$:
\begin{equation}
\theta^\dagger = \arg\min_{\theta} \frac{1}{n_{tr}} \sum_{i=1}^{n_{tr}} \sum_{x \in \mathcal{X}_i} \|u_i(x) - \mathcal{G}_\theta(a_i)(x)\|^2.
\end{equation}

We implement the neural operator architecture $\mathcal{G}_\theta = \mathcal{D}_\text{u} \circ \mathcal{E}_\text{a}$ where:
\begin{itemize}
    \item The encoder $\mathcal{E}_\text{a}$ maps geometric inputs $a_{\text{geom}}$ to spatially-grounded representations $z_a$.
    \item The decoder $\mathcal{D}_\text{u}$ combines these local geometric features with global parameters $\mu$ to produce output fields $u$.
\end{itemize}

\subsection{Model Architecture and Training}
Our architecture uses two main components: $f_{\theta_a, z_a}: \mathbb{R}^d \to \mathbb{R}^{n_a}$, $f_{\theta_u, z_u}: \mathbb{R}^d \to \mathbb{R}^{n_u}$ for input and output functions respectively. Both are implemented as conditional neural field architectures, where $\theta_a,\theta_u$ are global shared parameters and $z_a,z_u$ are sample-specific latent point cloud representations:
$$z = \{(p_j, c_j )\}_{j=1,..,N_{lat}}$$

where latent features $c_j\in \mathbb{R}^{l_{d}}$ are localized at spatial positions $p_j \in \mathbb{R}^d$ within the physical domain.

Input functions $a_i$, typically representing the geometry (such as SDF fields), are encoded into latent representations using the input encoder $z_{a_i} = \mathcal{E}_\text{a}({a_i})$ via meta-learning \cite{dupont2022data}, detailed below. Obtained local input representations are then processed with global parameters $\mu$ (when present), and decoded to output physical fields through the output spatial decoder $\mathcal{D}_u$: 
\begin{equation}
    \hat{u}_i(x) = \mathcal{D}_u(z_{u_{i}})(x) =f_{\theta_u,z_{u_{i}}}(x), \quad z_{u_i} = \{(p_j, [c_j; \mu])\}_{j=1}^{N_{lat}}. 
\end{equation}

Training proceeds by minimizing reconstruction losses for both input and output fields, as illustrated in Figure \ref{fig:improved_architecture}:

\begin{equation}
    \mathcal{L}^a = \sum_{i=1}^{n_{tr}} \frac{1}{|\mathcal{X}_i|} \sum_{x \in \mathcal{X}_i} \|a_i(x) - \mathcal{D}_a(\mathcal{E}_a(a_i))(x)\|^2,\quad \mathcal{L}^u = \sum_{i=1}^{n_{tr}} \frac{1}{|\mathcal{X}_i|} \sum_{x \in \mathcal{X}_i} \|u_i(x) - \mathcal{D}_u(z_{u_i})(x)\|^2
\end{equation}

where $\mathcal{D}_a$ represents the input decoder that reconstructs input functions from their latent representations.

\paragraph{Input Encoder}

Our encoder learns spatially anchored latent points that reconstruct the input function (geometry) via equivariant cross attention \cite{knigge2024space}:

\begin{equation}
    f_{\theta_a, z_a}(x_m) = W_o \sum_{j=1}^{N_{lat}} \text{att}_{m,j} \cdot v(b_{j,m}, c_j), \quad W_o \in \mathbb{R}^{n_a \times d_v}
    \label{eq:eq_dec}
\end{equation}

where $b_{j,m} = \gamma( x_m - p_j)$ represents the vector offset between query point $x_m$ and latent position $p_j$, composed with a  Fourier feature encoding layer $\gamma$, used to overcome spectral bias and capture high frequency components \cite{tancik2020fourier}. The attention weights incorporate both learned affinities and spatial locality as:

\begin{equation}
    \text{att}_{m,j} = \frac{\exp\left(\frac{q(b_{j,m})^T k(c_j)}{\sqrt{d_k}} - \sigma \|x_m - p_j \|^2 \right)}{\sum_{\ell=1}^{N_{lat}} \exp\left(\frac{q(b_{\ell,m})^T k(c_\ell)}{\sqrt{d_k}} - \sigma \|x_m - p_\ell \|^2 \right)}
    \label{eq:eq_att}
\end{equation}

where the Gaussian penalty term (controlled by window parameter $\sigma$) enhances local conditioning by penalizing distance from latent positions. The query and key functions are parameterized as:
$q(b_{j,m}) = W_q b_{j,m}, k(c_j) = W_k c_j$ $( W_q \in \mathbb{R}^{d_k \times d_\gamma}, W_k \in \mathbb{R}^{d_k \times l_d})$, and the value function combines relative position information with latent features through scale and shift modulation:

\begin{equation}
    v(b_{j,m}, c_j) = (W_v c_j) \odot (W_s b_{j,m}) + W_b b_{j,m}
    \label{eq:value_func}
\end{equation}
where $W_v \in \mathbb{R}^{d_v \times l_d}, W_s,W_b \in \mathbb{R}^{d_v \times d_{\gamma}}$.
For each input function $a_i$, the encoding follows a meta-learning optimization procedure based on CAVIA \cite{zintgraf2019fast,dupont2022data}. Latent features $c$ ($\forall j\in(1,N_{lat})$) are optimized through
$K=3$ gradient steps with learning rate $\alpha$:

\begin{align}
    c_j^{(0)} &= 0 \quad p_j^{(k)}=p_j^{(0)}=p_j\\
    c_j^{(k+1)} &= c_j^{(k)} - \alpha \nabla_{c_j^{(k)}} \mathcal{L}^a(f_{\theta_a,c_j^{(k)}}, a_i) \quad \text{for } 0 \leq k \leq K-1
\end{align}
this defines the encoder $\mathcal{E}_a(a_i) = z_{a_i}$ as an optimization procedure.
The latent positions $p_j$ are initialized to uniformly cover the input domain and remain fixed throughout optimization, as updating their values results in less stable training. This design choice prioritizes training stability while maintaining the spatial grounding essential for geometric representation.

During training, this inner loop optimization is performed in parallel with the global parameters $\theta_a$ optimization (outer loop). At inference, the global parameters are fixed, and encoding a new geometry requires only running the inner loop for a few gradient descent steps ($K$).

\paragraph{Output Decoder}

The output decoder $\mathcal{D}_\text{u}$ operates on the concatenation of input latent representations $c^{a_i}$ and global parameters (when present) $ z_{u_i} = \{(p_j, [c_j^{a_i}; \mu])\}_{j=1}^{N_{lat}}$ through two sequential operations: (i) latent self-attention and (ii) equivariant decoding.

The first operation enables communication between latent features, allowing mixing of global and local information. This key architectural feature operates on the latent features  :
$\tilde c_j = c_j + \operatorname*{SelfAttn}_j(\{c_\ell\}_{\ell=1}^{N_\text{lat}}).$
The self-attention mechanism is implemented as:

$$ \operatorname*{SelfAttn}_j(\{c_\ell\}) \;=\;
\sum_{\ell=1}^{N_\text{lat}}
   \operatorname*{softmax}_{\ell}
   \Bigl(
     \tfrac{(W_q c_j)^\top (W_k c_\ell)}{\sqrt{d_k}}
   \Bigr)
   (W_v c_\ell).
$$

This communication operation enables global feature mixing with computational complexity $O(N_{lat}^2)$, which is typically much smaller than the spatial mesh resolution, allowing the model to capture long-range dependencies and non-local physical phenomena without the computational overhead of full spatial attention.

Following feature communication, the enhanced latent representations $\tilde{z}_{u_i} = \{(p_j, \tilde{c}_j)\}_{j=1}^{N_{lat}}$ undergo equivariant decoding using the same translation equivariant architecture as the input encoder. The output reconstruction follows the same formulation as Equation \ref{eq:eq_dec}, with enhanced features:

$$
\hat{u}_i(x) = f_{\theta_u, \tilde{z}_{u_i}}(x) = W_o'\sum_{j=1}^{N_{lat}} \text{att}_{x,j} \cdot v'(b_{x,j}, \tilde{c}_j), \quad W_o' \in \mathbb{R}^{n_u \times d_v}$$

where the attention weights and value function follow the same parameterization as Equations \ref{eq:eq_att}-\ref{eq:value_func}, substituting $\tilde{c}_j$ for the enhanced latent features. The relative position encoding $b_{x,j} = \gamma(x - p_j)$ maintains translation equivariance, while the spatial positions $p_j$ remain unchanged to preserve geometric grounding.

\section{Experiments}
We experimentally demonstrate the validity of the proposed approach on two steady state PDE benchmarks: (i) the hyper-elastic material dataset and (ii) the AirfRANS Dataset, consisting of numerical solutions of the incompressible Navier-Stokes equations on 2D airfoils. Additionally, we provide an experiment on learning more complex geometry representations of multi-element airfoils in the Appendix \ref{app:multi_element}. The implementations and the code to reproduce the experiments can be found in the dedicated repository: \url{https://github.com/giovannicatalani/enf2enf_pytorch}.

\paragraph{The Hyper-elastic Material Dataset}
We consider a hyper-elastic material benchmark problem commonly used for geometry aware inference \cite{li2023fourier}. The boundary-value problem is defined on a unit cell domain, featuring an arbitrarily shaped void at its center. The underlying finite element solver employs approximately 100 quadrilateral elements. The input configuration is provided as a point cloud of around 1000 points representing the geometry, and the target output is the corresponding stress field. The dataset comprises 1000 training samples and 200 test samples. Although not representing a large scale PDE benchmark, this is a good demonstrative example that the proposed model has competitive performance beyond fluid dynamics applications. More details about the dataset are given in the Supplementary Materials.
Table~\ref{tab:model_comparison} reports the L2 mean relative errors on the hyper-elastic material dataset. The results indicate that the performance of FNO and UNet is limited, likely due to the interpolation steps required to map between unstructured and uniform grids. The two neural field based approaches, CORAL and the proposed \texttt{enf2enf} achieve the lowest relative errors, likely due to the explicit geometric encoding which is crucial for this task.

\begin{table}[htbp]
    \centering
    \caption{Comparison of Different Models. We show results on the AirfRANS Dataset (left) with Mean Squared Error (MSE) on the normalized predictions of the pressure field on the volume and on the surface, as well as the MSE error and the Spearman Correlation Coefficient $\rho_L$ on the lift coefficient predictions; and on the Hyper-elastic Material Dataset (right) with Mean L2 Relative Errors on the test set. Bold indicates best, underlined indicates second best.}
    \label{tab:model_comparison}
    \begin{minipage}{0.58\textwidth}
        \centering
        \begin{tabular}{lcccc}
            \toprule
            \multirow{2}{*}{Model} & \multicolumn{4}{c}{AirfRANS} \\
            \cmidrule{2-5}
             & Volume $\downarrow$ & Surface $\downarrow$ & $C_L$ $\downarrow$ & $\rho_L$ $\uparrow$ \\
            \midrule
            GraphSAGE      & 0.0087 & 0.0184 & 0.1476 & 0.9964 \\
            MeshGraphNet   & 0.0214 & 0.0387 & 0.2252 & 0.9945 \\
            Graph U-Net    & 0.0076 & 0.0144 & 0.1677 & 0.9949 \\
            GINO           & 0.0297 & 0.0482 & 0.1821 & 0.9958 \\
            Transolver & 0.0037 & 0.0142 & \underline{0.1030} & \underline{0.9978} \\
            CORAL & \underline{0.0035} & \underline{0.0120} & 0.1591 & 0.9964 \\
            \texttt{enf2enf}   & \textbf{0.0011} & \textbf{0.0032} & \textbf{0.0883} & \textbf{0.9989} \\
            \bottomrule
        \end{tabular}
    \end{minipage}%
    \hfill
    \begin{minipage}{0.38\textwidth}
        \centering
        \begin{tabular}{lc}
            \toprule
            \multirow{2}{*}{Model} & \multicolumn{1}{c}{Hyper-elastic} \\
            \cmidrule{2-2}
            & Test Error $\downarrow$ \\
            \midrule
            FNO            & 4.95e-2 \\
            UNet           & 5.34e-2 \\
            GraphNO        & 1.27e-1 \\
            DeepOnet       & 9.65e-2 \\
            Geo-FNO        & 3.41e-2 \\
            CORAL & \textbf{1.67e-2} \\
            
            \texttt{enf2enf}  & \underline{1.88e-2} \\
            \bottomrule
        \end{tabular}
    \end{minipage}
\end{table}

\begin{figure}[t]
    \centering
    \includegraphics[width=\textwidth]{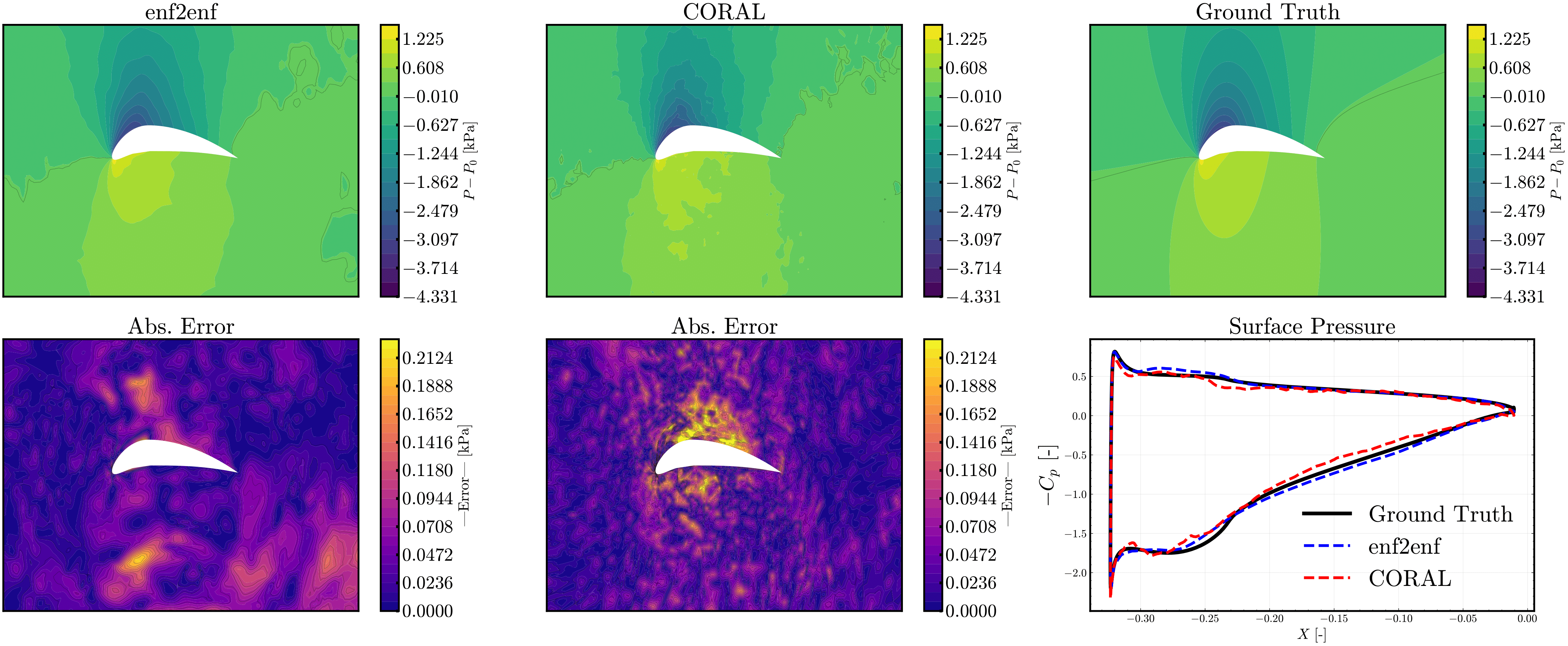}
    \caption{Visualization of pressure field predictions on an airfoil geometry. Top row: predictions from \texttt{enf2enf} (left) and \texttt{CORAL} (middle) compared to the ground truth (right). Bottom row: corresponding absolute errors for each model and surface pressure distribution. }
    \label{fig:pressure_comparison}
\end{figure}

\paragraph{The AirfRANS dataset}
The AirfRANS dataset \cite{bonnet2022airfrans} provides a challenging benchmark for surrogate modeling in computational fluid dynamics \cite{yagoubi2025neurips2024ml4cfdcompetition}, simulating the incompressible Reynolds-averaged Navier-Stokes (RANS) equations over a wide range of airfoil geometries and flow conditions. The geometries are systematically obtained through variations of the NACA 4- and 5-digit series airfoils, resulting in a diverse design space. The implicit distance (or SDF) field is defined over the entire mesh for each simulation as a continuous descriptor of the airfoil shapes. The flow conditions cover incompressible regimes with angles of attack ranging from -5° to 15°, representing typical subsonic flight conditions.  The dataset features a large scale PDE problem of industrial relevance, discretized on high resolution C-grid meshes with approximately 200,000 points per geometry, refined towards the airfoil surface to capture boundary layer effects. 
We focus on the \textit{full} task as 800 samples are used for training and 200 test samples for model evaluations. More details about the AirfRANS Dataset are given in the Supplementary Materials.
We mainly evaluate the model performance on the volumetric and surfacic pressure field predictions and the resulting lift coefficients, as these are the primary quantities of interest for aerodynamic design. \\
% LaTeX code for the revised paragraph
In Table \ref{tab:model_comparison}, it can be observed that \texttt{enf2enf} achieves the lowest errors across volumetric and surface pressure predictions as well as the lift coefficient estimation. The performance gap becomes more evident, especially on the surface pressure, as more complex flow features are localized close to the boundary, compared to the rest of the domain where the flow can be considered quasi-potential.  
Graph Neural Network (GNN) baselines such as GraphSAGE and Graph U-Net tend to perform better than MeshGraphNet for PDEs on large meshes thanks to the use of pooling and aggregation operations over extensive neighborhoods. Similarly to Graph U-Net, the proposed \texttt{enf2enf}, performs feature encoding on a latent set of nodes. However, Graph U-Net relies on complex pooling operations and aggregation steps to produce its latent representation, which is ultimately determined by the discretization of the input graph. Instead, \texttt{enf2enf} as a neural operator approach, yields representations that are independent of the specific input mesh instance, and can be decoded continuously in the spatial domain, thus being better suited to handle such tasks. \\
To assess the impact of explicit local feature encoding on capturing relevant aerodynamic phenomena, we next compare with global latent encoding methods such as CORAL in Figure \ref{fig:pressure_comparison}.
This aspect is especially crucial in fluid dynamics applications, where localized phenomena such as the leading edge suction peak play a critical role in determining aerodynamic performance. As depicted in Figure~\ref{fig:pressure_comparison}, the surface $C_p$ plot reveals that the maximum error at the leading edge is mitigated in the \texttt{enf2enf} predictions. Moreover, the isobaric lines outputs match more closely the high fidelity simulation with lower levels of noise (overfitting) and improved physical compliance.  

\paragraph{Discretization Convergence} Furthermore, we evaluate discretization convergence on the AirfRANS dataset by training each model at lower mesh resolutions (5k, 10k, 32k) and then testing on the full-scale meshes consisting of approximately 200k points. For direct inference, the entire high resolution mesh is processed in a single forward pass. For iterative inference, the mesh is divided into multiple subsamples, each processed in a separate forward pass until all 200k nodes are covered.
Our results show that graph based methods (GraphSAGE and GUNet) perform poorly under direct inference when the test resolution greatly exceeds the training resolution, although iterative inference can partially mitigate this at the cost of multiple runs. For instance, GraphSAGE at 32k training resolution requires roughly six passes to cover a 200k test mesh, significantly increasing total inference time. By contrast, Transolver can handle zero-shot super-resolution training at 5k and testing at 200k, yet sees degraded accuracy when the resolution gap grows. \texttt{enf2enf} remains discretization invariant and shows no difference between direct and iterative inference; it achieves high accuracy even when trained at much lower resolutions, thus avoiding the cost of large scale training while maintaining excellent performance at full resolution. All models achieve a remarkable 4 to 5 orders of magnitude speedup at inference compared to high fidelity CFD.

\begin{figure}[t]
    \centering
    \includegraphics[width=\textwidth]{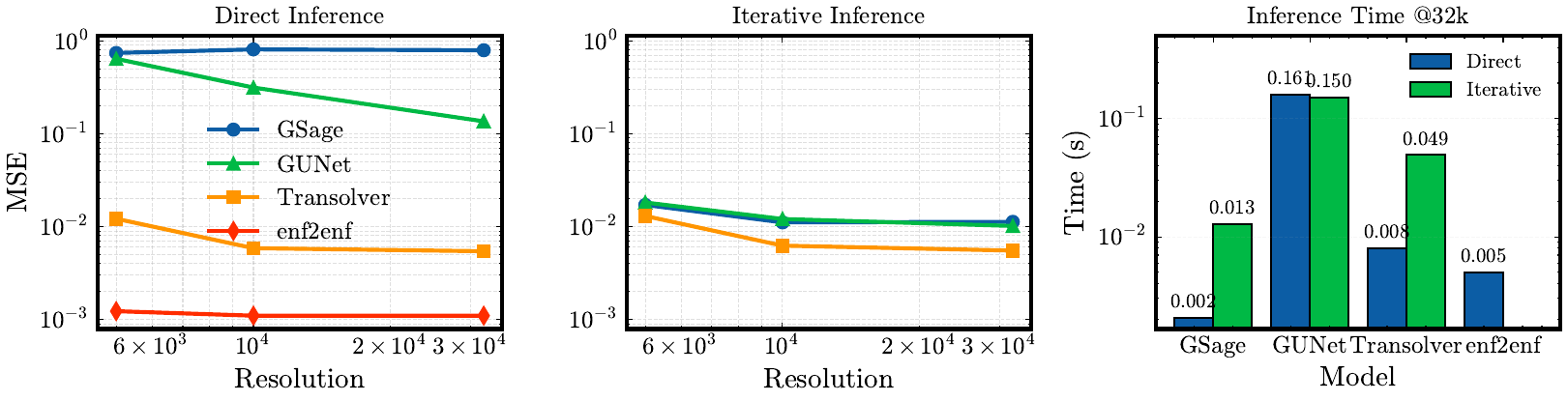}
    \caption{Discretization convergence analysis on the AirfRANS dataset. \textbf{Left:} Direct inference performance vs \textbf{training} resolution \textbf{Center:} Iterative inference; note that \texttt{enf2enf} is omitted in this panel due to its coordinate-based design. \textbf{Right:} Inference time histogram for models trained at a resolution of 32k points.}
    \label{fig:resolution_comparison}
\end{figure}

\section{Conclusion}

We introduced a geometry-aware operator learning framework leveraging spatially grounded neural fields within an encoder-decoder architecture for steady-state PDEs on complex geometries. Unlike approaches that incorporate explicit physics constraints in loss functions (e.g., Physics Informed Neural Networks (PINNs) \cite{raissi2019physics}), our method restricts the search space to functions respecting underlying PDE symmetries through translation equivariance, providing an alternative pathway for incorporating physical constraints.
Our approach leverages the discretization invariance of neural fields, beneficial for large-scale PDE modeling where training on full-resolution meshes is computationally prohibitive. 
Experimental validation on aerodynamic (AirfRANS) and material modeling benchmarks demonstrates competitive performance compared to graph-based methods, recent transformer models, and alternative neural field techniques, particularly for surface phenomena where local geometric features are critical.
Extension to three-dimensional problems requires investigation, likely necessitating more latent points and introducing computational complexity challenges.

\section*{Aknowledgements}
The experiments presented in this paper have been run on the ISAE-Supaero HPC cluster PANDO. 
This work was supported by Agence Nationale de la Recherche (ANRT), through  CIFRE PhD Fellowship sponsored by Airbus Operations SAS and ISAE-Supaero. 

\bibliographystyle{unsrt}
\bibliography{neurips_conference.bib}

\appendix

\section{Additional Experiment: Multi-element Airfoil}
\label{app:multi_element}
To further demonstrate the benefits of localized geometric encoding, we conducted experiments on multi-element airfoil shape reconstruction. We generated 200 geometric variations of the NASA GA(W)-I airfoil with a Fowler flap by systematically translating and rotating the flap while keeping the main airfoil fixed, representing typical high-lift device operations.
We compare two encoding approaches for SDF field reconstruction:
\begin{itemize}
\item[$\bullet$] Global encoding (CORAL/Functa): Uses a 32-dimensional global latent vector.
\item[$\bullet$] Local encoding (\texttt{enf2enf}): Uses 4 spatially-anchored points with 8-dimensional features each.
\end{itemize}

\begin{figure}[h]
   \centering
   \begin{tikzpicture}
       \node[inner sep=0] (img) {\includegraphics[width=0.9\textwidth]{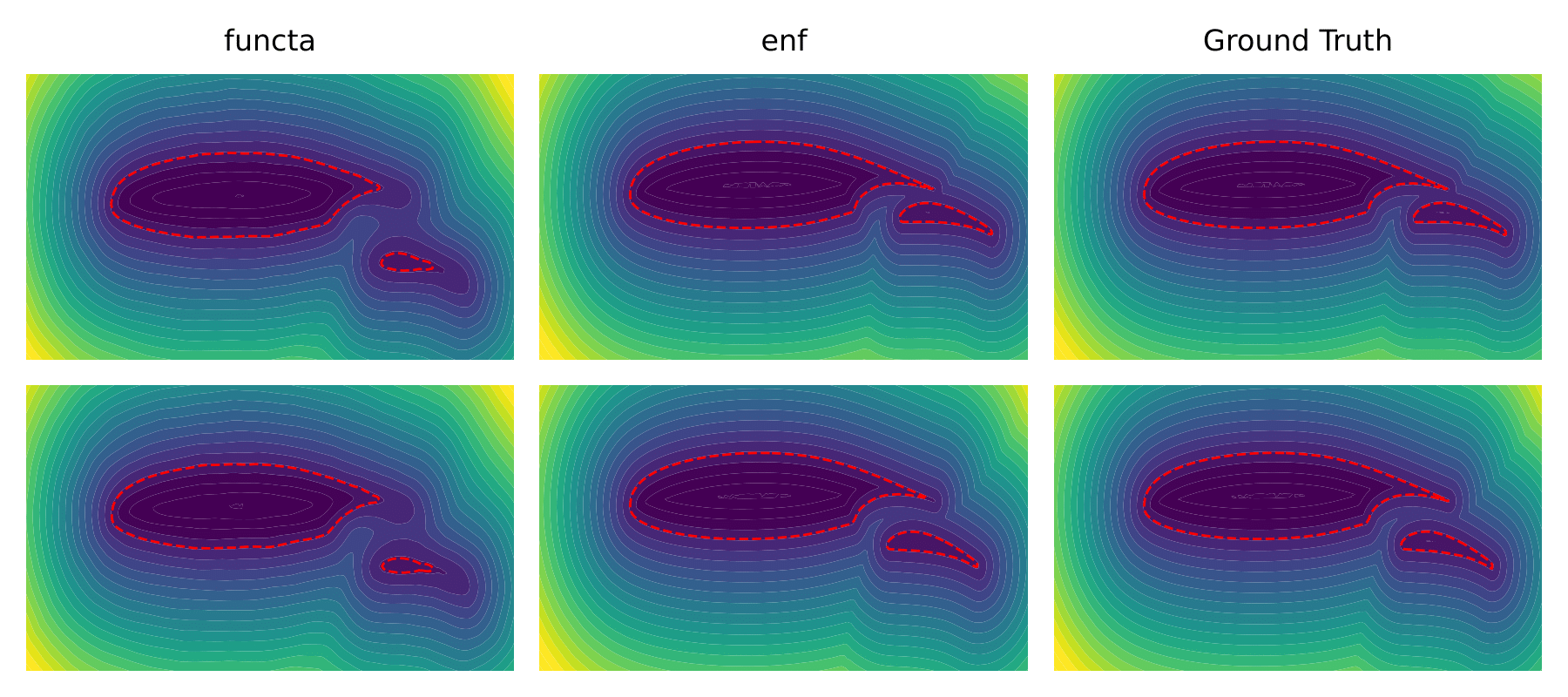}};
       
   \end{tikzpicture}
   \caption{Shape reconstruction results on the multi-element airfoil geometry. Red Line indicates the 0-level Signed Distance Function, corresponding to the airfoil shape. Global latents of shape 32 are used. Local latent point-clouds of 4 nodes and 8-dimensional node features are used. First row: high-lift configuration. Second row: high-lift and high-drag configuration.}
   \label{fig:multi_element_comparison}
\end{figure}

Dramatic improvement in reconstruction accuracy (3 orders of magnitude) are obtained using the local encoder: while global encoding treats each geometric variation as an entirely new shape to be encoded in its global latent space, the translation-equivariant architecture naturally handles the relative positioning of elements through its local feature encoding. This allows the model to effectively decompose the geometry into its constituent elements and capture their spatial relationships.
The qualitative results in Figure \ref{fig:multi_element_comparison} display how the local encoder accurately reconstructs both the main airfoil and flap geometries across different configurations, maintaining sharp features and precise relative positioning. 

\section{Implementation Details}
The implementation of our model architecture, training scripts, and dataset processing pipelines is available at \url{https://github.com/giovannicatalani/enf2enf_pytorch}. This repository contains all the code necessary to reproduce our experiments. Our proposed \texttt{enf2enf} model is implemented in PyTorch. For the baseline, we generally use the existing codebases in Pytorch.
\label{app:implementation}
\subsection{\texttt{enf2enf}}
The \texttt{enf2enf} architecture consists of two main components: an encoder network ($f_{\theta_a}$) that learns latent representations of input geometries, and a decoder network ($f_{\theta_u}$) that maps these representations to physical fields. Both components utilize Equivariant Neural Fields (ENFs) with translation equivariance as the core building block. The encoder follows a meta-learning optimization procedure based on CAVIA \cite{zintgraf2019fast}, where for each input function, the latent features $c_i$ are optimized through $K$ inner steps with learning rate $\lambda_c$, while the network parameters $\theta_a$ are updated through standard gradient descent.
Each ENF employs a multi-head attention mechanism where query points attend to a set of latent points $z := {(p_i, c_i)}_{i=1,..,N_{lat}}$. The attention scores are computed using both positional information and feature embeddings, with a Gaussian window (of size $\sigma$) function promoting locality. Random Fourier Features (RFF) are used to encode coordinate information, where inputs are projected using fourier features following $\gamma(x) = [\cos(Wx), \sin(Wx)]$ where $W \sim \mathcal{N}(0, \sigma_{RFF}^2)^{d \times d_{in}}$. The latent positions $p_i$ are initialized to uniformly cover the input domain and remain fixed during training to ensure stability.

\begin{table}[h]
   \centering
   \caption{Model Architecture and Training Specifications}
   \label{tab:network_specs}
   \begin{tabular}{lccc}
       \toprule
       & Hyper-parameter & AirfRANS & Elasticity \\
       \cmidrule{1-4}
       \multirow{7}{*}{\centering $f_{\theta_\text{a}}$ / $f_{\theta_\text{u}}$} 
       & Learnable Parameters & $\theta_\text{a},\{c_i\}_{i=1}^{n_\text{lat}}$ / $\theta_\text{u}$ & $\theta_\text{a},\{c_i\}_{i=1}^{n_\text{lat}}$ / $\theta_\text{u}$ \\
       & Gaussian Window $\sigma$ & 0.1 & 0.1 \\
       & Attention Heads & 2 & 2 \\
       & RFF $(d, \sigma_{RFF})$ & $(128, 1)$ / $(256, 2)$ & $(128, 2)$ / $(256, 10)$ \\
       & Latent Dims $(n_\text{lat}, l_\text{dim})$ & $(9, 8)$ / $(9, 8+2)$ & $(9, 8)$ / $(9, 8)$ \\
       & Network Width & 128 / 256 & 128 / 256 \\
       & N. Attention Blocks & 0 / 2 & 0 / 2 \\
       \cmidrule{1-4}
       \multirow{5}{*}{\centering Optimization} 
       & LR $\lambda_{\theta}$ & 1e-4 / 1e-4 & 1e-4 / 1e-4 \\
       & Inner Steps $K$ & 3 / - & 3 / - \\
       & Inner LR $\lambda_c$ & 1.0 / - & 0.5 / - \\
       & Downsampling & 10k / 10k & Full mesh \\
       & Epochs & 800 / 2500 & 1500 / 2500 \\
       \bottomrule
   \end{tabular}
\end{table}

\subsubsection{AirfRANS experiments}
For the AirfRANS dataset, we initialize the latent positions in a bounding box $[-0.5,0.25] \times [-0.25,0.25]$ closer to the airfoil surface, as most of the interesting flow phenomena occur in this region. The latent feature dimension in the decoder is increased from 8 to 10 to accommodate the concatenation of inflow conditions (velocity components) that globally condition the output field generation. The dataset is normalized by standardizing the output pressure fields and scaling the input coordinates to $[-1,1]$ through min-max normalization. Following the 'full' task setup, we use 800 samples for training and 200 for testing. Due to the large mesh sizes ($\sim 200,000$ points), we dynamically downsample to 10,000 points during both encoder and decoder training. At inference time, the continuous nature of Neural Fields allows us to evaluate on the full resolution meshes.
\subsubsection{Elasticity experiments}
For the elasticity problem, latent positions are initialized in a larger bounding box $[-0.75,0.75] \times [-0.75,0.75]$ to better cover the unit cell domain. The inputs and outputs are normalized following the same strategy as AirfRANS. Given the relatively small mesh sizes ($\sim 1000$ points), we can process the entire point cloud during training without downsampling, leading to better reconstruction of sharp features near the void boundaries. 
For each input function, we compute a geometry descriptor by encoding the displacement field between the average mesh position and the current sample's points, as done in previous works \cite{serrano2024operator}. 

\subsection{Baseline Models}
We compare our approach against several state-of-the-art methodologies across different architectural paradigms:
\paragraph{Neural Field Baselines}
We implement  the CORAL architecture, modified to use Random Fourier Feature encoding instead of the original SIREN networks. To ensure a fair comparison, we employ a 64-dimensional latent space, which approximately matches our model's representational capacity (9 latent points × 8 features). The Fourier Feature encoding parameters are kept consistent with our approach to isolate the impact of the architectural differences.
\paragraph{Graph-Based Approaches}
We evaluate against the graph neural network baselines presented in the AirfRANS benchmark, including GraphSAGE, MeshGraphNet, and Graph U-Net. For computational efficiency on the AirfRANS dataset, these models operate on downsampled meshes of 32,000 points while maintaining the original mesh connectivity structure. This sampling strategy provides a balance between computational feasibility and preservation of important flow features.
\paragraph{Neural Operator Methods}
For the AirfRANS dataset, we compare against Transolver, maintaining their original implementation settings with input meshes downsampled to 32,000 points. On the elasticity benchmark, we include comparisons with Geo-FNO and traditional convolutional architectures (U-Net, FNO) as reported in previous studies. It's worth noting that the grid-based methods (U-Net, FNO) require an additional interpolation step to map between the unstructured mesh and regular grid representations, which can impact their performance on complex geometries.

\section{Datasets Details}
\label{app:datasets}

\subsection{AirfRANS Dataset}
The AirfRANS dataset consists of numerical solutions of the Reynolds-Averaged Navier-Stokes (RANS) \cite{pope2001turbulent} equations for incompressible flow around airfoil geometries \cite{bonnet2022airfrans}. The RANS equations are derived by applying Reynolds decomposition to the velocity and pressure fields:
\begin{equation}
u_i = \bar{u}_i + u_i', \quad p = \bar{p} + p'
\end{equation}
where $\bar{\cdot}$ denotes the ensemble-averaged quantity and $\cdot'$ represents the fluctuating component. Substituting these decompositions into the incompressible Navier-Stokes equations and taking the ensemble average leads to:
\begin{equation}
\partial_i \bar{u}_i = 0
\end{equation}
\begin{equation}
\partial_t \bar{u}_i + \partial_j(\bar{u}_i \bar{u}_j) = -\frac{1}{\rho}\partial_i \bar{p} + \nu \partial^2_{jj}\bar{u}_i - \partial_j(\overline{u'_i u'_j})
\end{equation}
The Reynolds stress tensor $\tau_{ij} = -\overline{u'_i u'_j}$ is modeled using the Boussinesq hypothesis, which introduces the concept of turbulent viscosity $\nu_t$:
\begin{equation}
\tau_{ij} = \nu_t \left(\partial_i \bar{u}_j + \partial_j \bar{u}_i\right) - \frac{2}{3}k\delta_{ij}
\end{equation}
where $k = \frac{1}{2}\overline{u'_i u'_i}$ is the turbulent kinetic energy. Incorporating this model and defining an effective pressure $\bar{p} := p + \frac{2}{3}\rho k$, the final form of the incompressible RANS equations becomes:
\begin{equation}
\partial_i \bar{u}_i = 0
\end{equation}
\begin{equation}
\partial_t \bar{u}_i + \partial_j(\bar{u}_i \bar{u}_j) = -\frac{1}{\rho}\partial_i \bar{p} + \partial_j[(\nu + \nu_t) \partial_j \bar{u}_i]
\end{equation}
In the AirfRANS dataset, these equations are solved using the Spalart-Allmaras (SA) turbulence model \cite{spalart1992one}, which introduces a single transport equation for a modified eddy viscosity $\tilde{\nu}$. The turbulent viscosity is then computed as $\nu_t = f_{v1}\tilde{\nu}$, where $f_{v1}$ is a damping function that ensures proper near-wall behavior. The SA model is particularly well-suited for aerodynamic applications as it was specifically calibrated for aerodynamic flows with adverse pressure gradients and separation.
The dataset provides numerical solutions obtained using OpenFOAM's \cite{jasak2009openfoam} incompressible solver with the SA turbulence model. The simulations are performed on high-resolution C-grid meshes ($\sim$200,000 cells) with appropriate refinement near the airfoil surface to resolve the boundary layer ($y^+ < 1$). The flow conditions span Reynolds numbers from 2 to 6 million, with angles of attack ranging from -5° to 15°. The entire dataset generation process, validation and benchmarking is described in the original paper \cite{bonnet2022airfrans}.

\subsection{Hyper-elastic Material Dataset}
The structural mechanics problem considered in this work involves the analysis of a hyperelastic material described by the momentum conservation equation:
\begin{equation}
\rho_s \frac{\partial^2 u}{\partial t^2} + \nabla \cdot \sigma = 0
\end{equation}
where $\rho_s$ represents the material density, $u$ denotes the displacement field, and $\sigma$ is the Cauchy stress tensor. The problem domain consists of a unit square $\Omega = [0, 1] \times [0, 1]$ containing a centrally located void with variable geometry. To generate physically meaningful void configurations, a stochastic process for the void radius is defined as:
\begin{equation}
r = 0.2 + \frac{0.2}{1 + \exp(\tilde{r})}, \quad \tilde{r} \sim \mathcal{N}(0, 4^2(-\nabla + 32)^{-1})
\end{equation}
This formulation naturally enforces the physical constraint $0.2 \leq r \leq 0.4$ on the void radius.
The material behavior is characterized by an incompressible Rivlin-Saunders constitutive model, setting the material parameters to $C_1 = 1.863 \times 10^5$ and $C_2 = 9.79 \times 10^3$, representing a soft rubber-like material. The boundary conditions consist of: fixed displacement (clamped) condition on the bottom edge, traction force $t = [0, 100]$ and traction-free conditions on all other boundaries.
The dataset was generated using a finite element solver employing approximately 100 quadratic quadrilateral elements, with each simulation requiring around 5 CPU seconds to complete. The input geometry is represented by point clouds containing roughly 1000 points that describe the void shape and domain boundaries. The target output fields consist of the stress components computed at these same locations. The dataset split used in this paper experiments comprises 1000 training samples and 200 test cases, following the settings of the orginal work introducing the dataset \cite{li2023fourier}.

\section{Ablation Studies}

In this section, we conduct systematic ablation studies to validate key architectural design choices in our \texttt{enf2enf} model. We focus on two main aspects: (1) the impact of latent representation capacity through varying numbers of latent points and feature dimensions, and (2) the comparison between our direct encoder-decoder architecture versus traditional encoder-processor-decoder strategies. All ablation experiments are conducted on the AirfRANS dataset using consistent training procedures and evaluation metrics.

\subsection{Latent Representation Capacity}

We investigate how the latent representation capacity affects model performance by varying the number of latent points ($n_{\text{lat}}$) and latent feature dimensions ($l_{\text{dim}}$) while maintaining approximately constant total representational capacity. Table~\ref{tab:latent_ablation} presents the results for three configurations: sparse-high dimensional [4,16], balanced [9,8], and dense-low dimensional [16,4] representations.

\begin{table}[h]
    \centering
    \caption{Impact of Latent Representation Configuration on AirfRANS Performance. The Mean Square Error (MSE) on the test set for the Signed Distance function encoder reconstruction, and on the final output error is recorded.  }
    \label{tab:latent_ablation}
    \begin{tabular}{ccc}
        \toprule
        $n_{\text{lat}}$, $l_{\text{dim}}$  & MSE (SDF) & MSE (Output)\\
        \midrule
        $[4, 16]$  & 1.9e-6 & 4.06e-3\\
        $[9, 8]$  &1.8e-7 & 1.13e-3 \\
        $[16, 4]$  & 1.7e-7 & 2.61e-3 \\
        \bottomrule
    \end{tabular}
\end{table}

\begin{figure}[t]
    \centering
    
    % Create a row of titles
    \makebox[0.9\textwidth]{%
        \hspace{-0.1\textwidth}enf far-field%
        \hspace{0.17\textwidth}enf close%
        \hspace{0.20\textwidth}functa%
    }
    
    % Small space between titles and image
    \vspace{0.05cm}
    
    % The image
    \includegraphics[width=0.9\textwidth, trim={0 0 0 0.5cm}, clip]{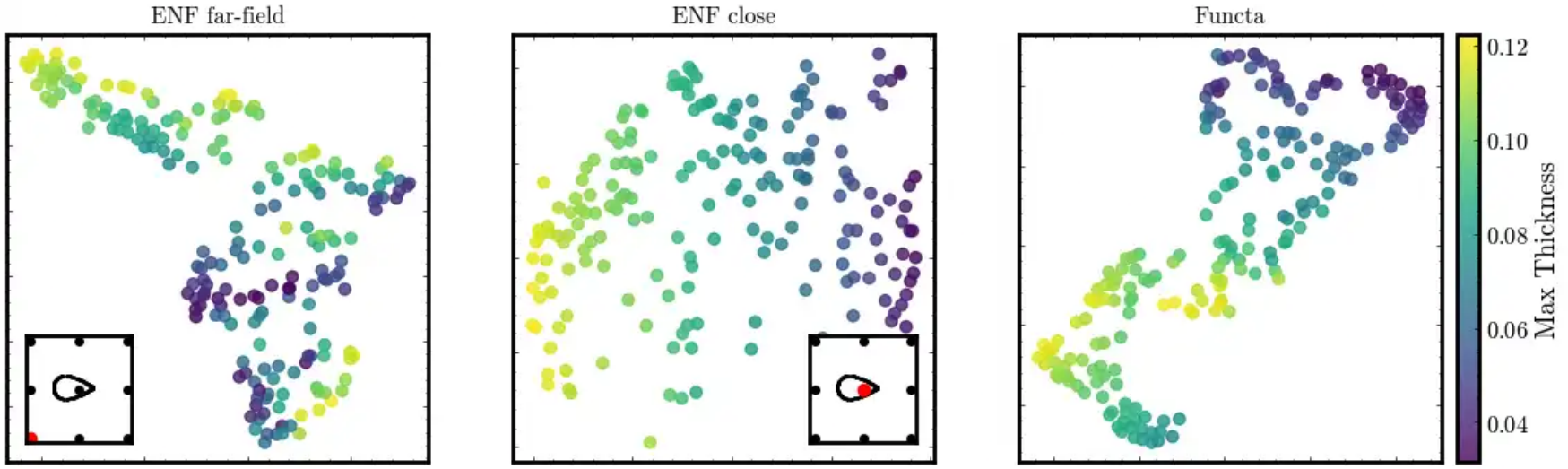}
    
    \caption{t-SNE visualizations of latents fitted on the SDFs of AirfRANS test split. \textbf{Left:} 8-dimensional latent code computed at a farfield point of the latent point cloud. \textbf{Center:} latent code at a central node located on the airfoil. \textbf{Right:} t-SNE visualization of the 64-dimensional global embedding obtained via functa.}
    \label{fig:latents_tsne}
\end{figure}

The results demonstrate an optimal trade-off exists between spatial resolution (number of latent points) and feature expressiveness (latent dimensions). A sufficient number of latent points allows to capture finer geometric details (thus achieving better input field reconstruction), which translate to improved downstream performance. Further increasing the number of latent points leads to more in the SDF reconstruction but worse downstream performance.
Finally, the t-SNE visualizations in Figure~\ref{fig:latents_tsne} further illustrate that \texttt{enf2enf} effectively learns and disentangles local geometric features, yielding a well structured latent space. The local latent codes computed at farfield and central nodes (left and center panels) exhibit clear clustering that correlates with key geometric parameters, such as airfoil thickness. Notably, the farfield latent feature (left panel) shows little sensitivity to thickness variations because its spatial position is distant from the airfoil, and therefore it does not contribute significantly to reconstructing the SDF field in regions where thickness is a critical factor. In contrast, the 64-dimensional array latent representation from \texttt{functa2functa} (right panel) is less structured, as it aggregates local features into a single global representation. Consequently, the localized encoding in \texttt{enf2enf} enables the decoder to focus on significant aerodynamic features while filtering out less relevant local variations.\\

\end{document}